\documentclass[letterpaper, 10 pt, conference]{ieeeconf}  

\IEEEoverridecommandlockouts                              

\usepackage{graphics} 
\usepackage{amsmath} 

\usepackage{xcolor}

\title{\LARGE \bf 
Direct and inverse modeling of soft robots by learning a condensed FEM model 
}

\author{Etienne Ménager$^{*}$, Tanguy Navez$^{*}$,  Olivier Goury$^{*}$ and Christian Duriez$^{*}$ 
\thanks{Etienne Ménager and Tanguy Navez contributed equally to this work.  $^{*}$Univ. Lille, Inria, CNRS, Centrale Lille, UMR 9189 CRIStAL, F-59000 Lille, France. 
        {\tt\small etienne.menager@inria.fr, \tt\small tanguy.navez@inria.fr}}%
}

\begin{document}

\maketitle
\thispagestyle{empty}
\pagestyle{empty}

\begin{abstract} 
The Finite Element Method (FEM) is a powerful modeling tool for predicting the behavior of soft robots. 
However, its use for control can be difficult for non-specialists of numerical computation: it requires an optimization of the computation to make it real-time. 
In this paper, we propose a learning-based approach to obtain a compact but sufficiently rich mechanical representation.
Our choice is based on nonlinear compliance data in the actuator/effector space provided by a condensation of the FEM model. 
We demonstrate that this compact model can be learned with a reasonable amount of data and, at the same time, be very efficient in terms of modeling, since we can deduce the direct and inverse kinematics of the robot.
We also show how to couple some models learned individually in particular on an example of a gripper composed of two soft fingers.
Other results are shown by comparing the inverse model derived from the full FEM model and the one from the compact learned version.
This work opens new perspectives, namely for the embedded control of soft robots, but also for their design. These perspectives are also discussed in the paper. 

\end{abstract}

\section{INTRODUCTION}
\label{INTRODUCTION}

This last decade, large advances in the modeling and control of soft robots have been achieved.
In the early 2010s, when the soft robotics community started to experience a very strong growth, it was considered that modeling and controlling soft robots was a challenge for which no general solution existed \cite{trivedi2008soft}\cite{lipson2014challenges}\cite{majidi2014soft}.
Since then, solutions have been proposed that can be classified into three categories: geometric methods, mechanical models and learning approaches.
Geometric methods are based on strong hypotheses concerning the behavior of the robot, in particular on hypotheses of constant or piecewise constant curvatures \cite{webster2010design}. They are therefore efficient for some types of robots, especially continuum robots, and for some applications, especially when the actuation is well distributed along the main axis of the robot.
The mechanical models are based on the mechanics of continuous media. Depending on the assumptions applicable to the robot, the work is based on the theory of beams (non-deformable section) of Euler- Bernoulli \cite{shapiro2015modeling} and Cosserat \cite{renda2018discrete}, or on the more general theories of 3D deformable solids, with in particular the use of FEM model \cite{duriez2013control}. These approaches are more general but require expertise in numerical modeling, especially to make the computation fast enough for robotic applications. 
Machine learning approaches have been applied to soft robots \cite{kim2021review} using the numerous methods and software libraries that have been developed in recent years. 
From input-output data, an algorithm learns and then predicts the robot behavior. 
Learning approaches sometimes use mechanical models to generate the needed data, but in this case, the model simulation is often used as a black box.
So far in the literature, mechanical and learning approaches are relatively distinct. 
In this paper, we propose, on the contrary, a "white box" approach where we take advantage of the structure of the mechanical model to perform targeted learning.

\textbf{Contribution:} This paper introduces a compact mechanical representation (MR) of a soft robot based on a FEM model.  
The compact representation is based on the data of a condensed compliance matrix of the robot in the actuator / effector spaces.
This matrix varies in a non-linear manner in the workspace of the robot, but we show that we can learn these data with a standard method, the MultiLayer Perceptron (MLP) \cite{MLP}. The method is evaluated on two FEM simulations of soft robots by testing the obtained inverse models in quasi-static configuration.
We also show how the individually learned models can be combined through the case of a gripper composed by 2 soft fingers which model has been learned individually. This method opens interesting perspectives because it enables to obtain a mechanical model usable for the control and which is differentiable from data obtained on a finite element model of a soft robot.

\section{BACKGROUND AND RELATED WORK}
\label{BACKGROUND}

\subsection{Soft robot quasi-statics behavior with Finite Element Method}
\label{model_and_simu}

The mechanical behavior of soft robot deformations can be described through continuum mechanics for which there are no analytical solutions in the general case. Non linear FEM  is one of the numerical methods employed to obtain a converging approximate solution. We focus on quasi-static formulation without contacts. Based on solid mechanics, the quasi-static equilibrium is written:

\begin{equation}
\begin{aligned}
     K(x) \Delta x = f_{ext} - f_{int}(x)
     + H_{a}^T \lambda_{a}
\end{aligned}
\label{eqQuasiStaticEquilibrium}
\end{equation}

where  $\Delta x$ are small displacement of nodes around their current position $x$, $K(x)$ the tangent stiffness matrix depending on the current positions of the FEM nodes, $f_{ext}$ are external forces such as gravity forces,  $f_{int}$ are non linear internal forces of the deformable structure computed from the material laws chosen for the robot. Actuation constraints $H_{a}^T \lambda_{a}$ are formulated as Lagrangian multipliers. This is convenient both for forward and inverse modeling. 

In the inverse problem, the constraints are not known in advance and we face quadratic optimization problems solved for the actuation with dedicated solvers \cite{inverse} \cite{trunk_inverse}. A soft robot motion is then controlled by specific points called effectors. The most common use is their control in position where the squared norm of the shift $\delta_{e} = x_{effector} - x_{goal}$ between effectors and goals is minimized. The resolution is done in two steps. First, a free configuration of the robot is solved without any constraint:
\begin{equation}
\begin{aligned}
     K(x) \Delta x^{free}_{i} = f_{ext} - f_{int}(x_{i-1})
\end{aligned}
\label{eqQuasiStaticFreeEquilibrium}
\end{equation}

After solving this system for $\Delta x^{free}$, it is then possible to evaluate $\delta_{e}^{free}$ which corresponds to the violation of the objective if there is no actuation. This violation is updated by a linear approximation of the nodes displacement using FEM interpolation. Using the Schur complements gives:
\begin{equation}
\begin{aligned}
    \delta_{e} = W_{ea} \lambda_a +  \delta_{e}^{free} \\
    \delta_{a} = W_{aa} \lambda_a + \delta_{a}^{free} \\
\end{aligned}
\label{eq:schur}
\end{equation}
where the $W_{ij} = H_i K^{-1}(x) H_j^T \text{ for } i,j \in {e,a}$ are the Schur complements \cite{qpoases}  i.e. the compliance matrices projected in constraint space. They gather the mechanical coupling between effectors $e$ and actuators $a$. The $\delta_i \text{ for } i \in {e,a}$ are homogeneous to distances and computed from the same distance vector $\delta_i^{free}$ evaluated at the previous equilibrium.

Solving the inverse problem finally amounts in solving the following optimisation problem: 
\begin{equation}
    \begin{aligned}
     arg_{\lambda_a}&min \quad ||\delta_{e}||^2 \\
    s.t. &  \delta_{min} \leq \delta_a \leq \delta_{max} \text{ (Actuators course constraint)}  \\ 
    &  \lambda_{min} \leq \lambda_a \leq \lambda_{max} \text{ (Actuation effort constraint)}  \\ 
    \end{aligned}
\label{eq:QPActuation}
\end{equation}

In the direct problem, when the actuator forces intensity $\Delta \lambda_{a}$ is varying then the corresponding motion of the effectors $\Delta \delta_{e} = W_{ea} \Delta \lambda_{a}$ can be computed. As actuators are also controlled in position, varying the position of the actuators $\Delta \delta_{a}$ lead to obtain the motion created on the effectors through $\Delta \delta_{e} = W_{ea} W_{aa}^{-1} \Delta \delta_{a}$. This equation provides the Jacobian $ J = W_{ea} W_{aa}^{-1}$ of the soft robot direct kinematics.

This background shows that both direct and inverse models of soft robots can be derived as long as an approximation of equations \ref{eq:schur} given the effective state of the actuation $\delta_{a}$ is available. In the case of a cable actuator, this value corresponds to the effective length of the cable. 
    
\subsection{State Representation Learning and robotics}
\label{srl}

State Representation Learning (SRL) algorithms \cite{srl_overview} are applied for learning small dimension features called states for fully describing an agent. These methods are widely used in robotics for obtaining a compact representation of robots when many descriptive sensors are involved. Small dimensional states are often used for solving control tasks using methods such as Reinforcement Learning. By keeping only the useful features, better time and computation efficiency are reached. This is a critical point for this kind of methods, since testing actions in the environment or retrieving state is expensive. 

Mathematically speaking, an agent is considered within an environment with which it can interacts using actions $a_t$ belonging to a given action space. Each action makes the agent go from a state $s_t$ to a state $s_{t+1}$, unknown but assumed to exist. The agent is not aware of his own state, but only of observations $o_t $ belonging to a given observation space. The idea behind SRL is to learn a small dimension representation $ \widetilde{s}_t$ of the state $s_t$ from previous observations. This amounts to find a mapping $\psi$ such that $\widetilde{s}_t = \psi(o_{1:t})$.

In machine learning and more particularly in the robotics and control fields many methods have been implemented \cite{srl_overview}: observation reconstruction such as Principal Component Analysis \cite{MOR}, learning a direct or inverse model for example using direct model and encoder \cite{bern_2020}, using prior knowledge in the cost function to constrain the state space like robotic prior \cite{robotic_prior}, or integrating relational biases in data \cite{robot_structure} \cite{robot_dynamic}. These methods introduce a compact reduced model of a robot and tools for learning or reconstructing it. Compared to these different end-to-end methods, our method enables to learn a mechanical model that is not specific to a single task.

\section{MATHEMATICAL MODELING}
\label{mathematical_modeling}

\subsection{Mechanical representation of soft robot}
\label{OurMR}

In this work, a high dimensional state of the robot $s_t$ is available through simulation. When using FEM methods, the state of the robot is the position of all the vertices of its mesh. For reducing the dimension of the robot state, the global compliance matrix $W$ projected in the constraint space (see section \ref{model_and_simu}) is a good candidate because it is a small dimensional matrix, independent of the size of the FEM mesh and it expresses a direct relationship between actuators and effectors. In combination with the free displacement vector $\delta^{free}$, it is relevant enough for modeling and controlling a soft robot. However, the computation of $W$ is expensive because it involves the multiplication and inversion of several large matrices and this needs to be computed at every simulation step. This is an incentive for learning it rather than computing it from mechanical simulation. Learning these mechanical quantities rather than non-supervised one makes it possible to obtain interpretive robot representation useful for simulating and controlling the robot using mechanical equations. 

The learning problem is then resumed as the following: given observations $o_t = (\delta_{a,t}, W_t, \delta^{free}_t)$ where $\delta_{a,t}$, $W_t$ and $\delta^{free}_t$ are respectively the effective state reached by the actuation constraint when applying a specific actuation and corresponding projected compliance matrix and free displacement of the robot, the goal is to reconstruct the sub-part of the observation $(W_t, \delta^{free}_t)$ from initial observation $o_0$ and current constraint state $\delta_{a,t}$. This leads to learn a function $F$ such that
\begin{equation}
\widetilde W_t, \widetilde \delta^{free}_t = F(\delta_{a,t}, W_0, \delta^{free}_0)
\label{global_function}
\end{equation}
where $W_0$ is the compliance matrix at time $t=0$. This is a good compliance initialization since it already captures the mechanical coupling between constraints and the compliance matrix varies smoothly from that initial position during simulation. This mathematical formulation exclude cases when the robot may have several configurations for a same actuation i.e. cases where the simulation falls into a local minimum.

\begin{figure*}[!ht]
\centering
\resizebox{.7\textwidth}{!}{
\includegraphics{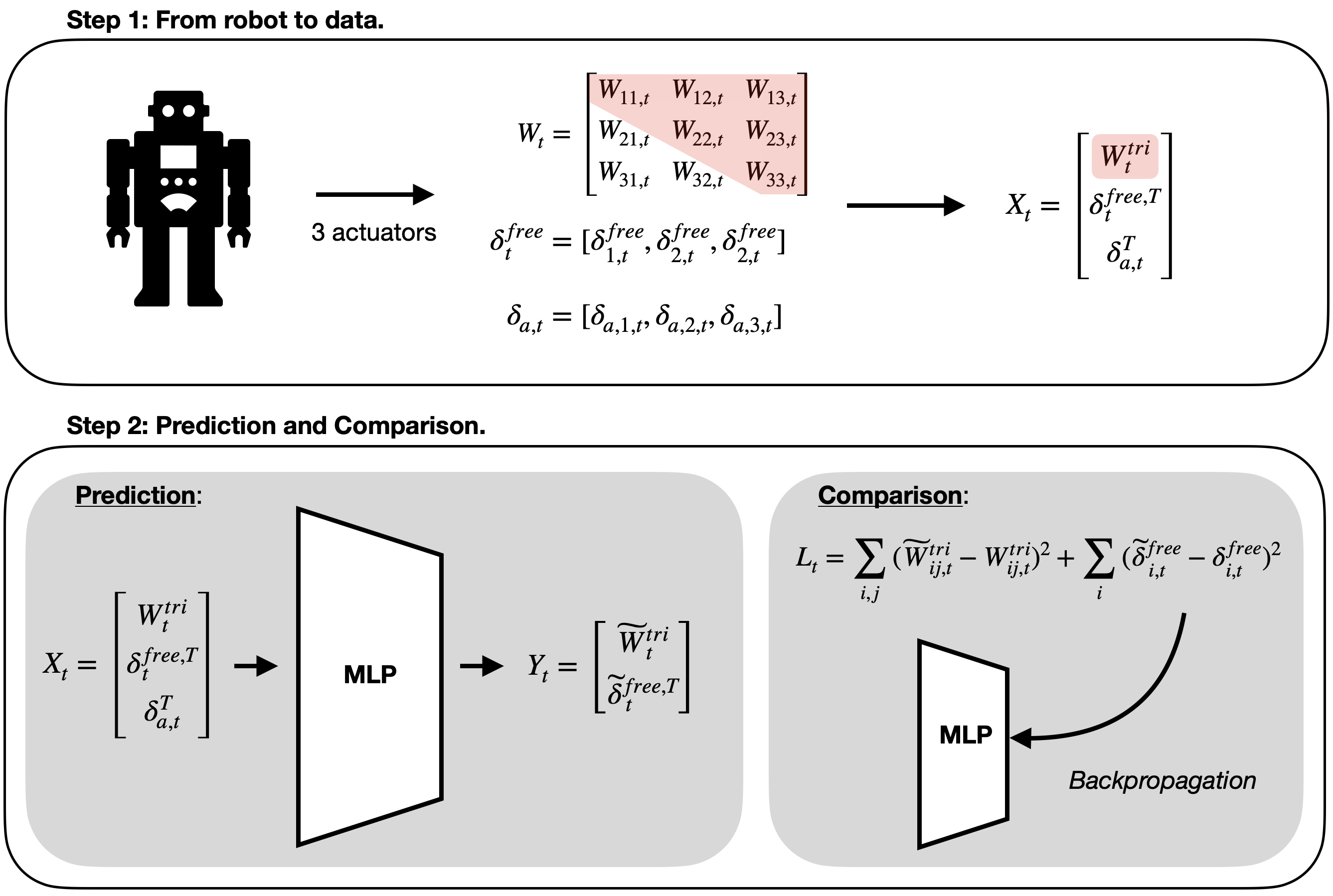}
}
\caption{Schematic of the different learning steps involved in the MR learning framework on the example of a robot with 3 actuator constraints. Step 1: Extraction of the robot constraints and creation of the data. Step 2: Prediction of $(\widetilde W^{tri}, \widetilde \delta_{t}^{free})$ and comparison between the prediction and the true value of $W^{tri}$ and $\delta^{free}$.}
\label{im_MLP}
\end{figure*}

\subsection{Learning models}
\label{learning_model}

As described in section \ref{model_and_simu}, both actuation and effector constraints are considered in this work. The learned compliance matrix $W$ is a symmetric matrix that represents the relations between the different constraints. It is formed by the Schur complements introduced in Eq. \ref{eq:schur}:
\begin{equation}
W_t = 
\begin{bmatrix}
W_{ee,t} & W_{ea,t}\\
W_{ae,t} & W_{aa,t}\\
\end{bmatrix}
\label{eq:W}
\end{equation}
The magnitude of both the values of $W_t$ and $d^{free}_t$ depends of the type of constraints and the measurement units chosen for the simulation. For avoiding vanishing gradients during the learning process, both the input and output data are element-wise standardized. This standardization is computed once using all the training set. The values of $W_t$ and $d^{free}_t$ related to the actuator space and those related to the effector space can be of different order of magnitude. In order not to overload the notations, we consider standardized inputs for learning and not standardized inputs for control afterward.

As the matrix is symmetric, the networks are trained in a supervised manner through minimizing the reconstruction loss 
\begin{equation}
L = \sum_t [||\widetilde W_t^{tri} - W_t^{tri}||^2 +  ||\widetilde  \delta_{a,t}^{free} - \delta_{a,t}^{free}||^2]
\label{eq:Loss}
\end{equation}
where $W_t^{tri}$ is the upper triangular part of the matrix $W_t$, $\widetilde W_t^{tri}$ and $\widetilde  \delta_{a,t}^{free}$ are the predicted values of $W_t^{tri}$ and $\delta_{a,t}^{free}$ at time $t$ given by the networks.

Although we also have considered a more complex architecture for the MR learning described in Section \ref{CONCLUSION}, a single MLP is sufficient for learning a MR for control tasks without considering contacts. A MLP is a type of neural network organized in several layers in which information flows from the input to the output layer only. In each layer, a learnable affine transformation of the input is followed by the applications of a non linear function according to
\begin{equation}
    X_{n+1} = \sigma_n (\omega_n X_n + b_n)
\label{eq:MLP}
\end{equation}
where $X_{n+1}$ is the output of the layer $n$, $\sigma$ a non linear activation function and $(\omega_n, b_n)$ the weight and bias of the layer $n$. In our implementation, $X_{0}$ is the raw data vectorized and concatenated to form a single vector, $\sigma$ is the $ReLu$ function and $(\omega_n, b_n)$ are the learnable weights of the network. The output $X_N$ where $N$ is the number of layers of the MLP is the vectorized couple $(W_t, \delta_{t}^{free})$. 

The MR learning framework is summarized in Fig. \ref{im_MLP}.

\subsection{Robot control pipeline from mechanical representation}
\label{pipeline}

The soft robot control problem is expressed as a quadratic optimization problem provided in Eq. \ref{eq:QPActuation}. The matrices $\widetilde W_{ea}$ and $\widetilde W_{aa}$ as well as the vector $\widetilde \delta_a^{free}$ and $\widetilde \delta_e^{free}$ are learned variables. The position of the goal does not depend on the configuration of the robot in the formulation of $\delta_e^{free}$. Even if $\widetilde \delta_e^{free}$ is learned for a specific goal position, this value can then be easily updated for other goals. For controlling a soft robot using MR, the complete pipeline therefore consists of an off-line learning phase followed by an online control algorithm applied on learned variables.

\subsubsection{Off-line learning}: As the framework is a supervised learning framework, pre-computation steps where data are collected and the learning model is trained are first performed. The data $(W^{tri}_t, \delta_t^{free}, \delta_{a,t})$ are obtained through simulation and stored in a database. Each data point is associated to a different robot configuration after applying an action $a_t$. The range of possible values of $a_t$ is bounded in a continuous interval. Sampling the actuation space is done using homogeneous grid search for building the training dataset and random search for the test dataset. The number of points of the test set is chosen to amount for $25\%$ of the number of points of the training set. Once the data is acquired, learning consists in training the network to predict the quantities $(W^{tri}_t, \delta_t^{free})$ from the quantities $(W^{tri}_0, \delta_0^{free},\delta_{a,t})$ using Eq. \ref{eq:MLP} and \ref{eq:Loss}.

\subsubsection{Online control}: The trained network is used for controlling the soft robot according to the following scheme:
\begin{enumerate}
         \item $\delta_{a,t}$, $W^{tri}_0$ and $\delta_0^{free}$ are recovered from FEM simulation. During the first iteration of a soft robot control experiment, $\delta_{a,t}$ is obtained for a zero actuation effort applied on the actuators.
         \item Prediction of $(\widetilde W^{tri}_t, \widetilde \delta_{a,t}^{free}) = F(W^{tri}_0, \delta_0^{free}, \delta_{a,t})$ using the learned network. $\widetilde W_t$ is reconstructed from $\widetilde W_t^{tri}$. In order to entirely reconstruct $\delta_{t}^{free}$ , $\delta_{e,t}^{free}$ is build from the distances between goals and effectors at time $t$. 
         \item The inverse optimization problem from Eq. \ref{eq:QPActuation} is solved using a quadratic optimization solver. The result is the actuation effort vector $\lambda_{a,t}$ for controlling the robot. In our implementation, the solver from the C++ library qpoases \cite{qpoases} is used, but others can be used. 
         \item The actuation effort $\lambda_{a,t}$ is applied in the simulation and a FEM time step is computed for updating the actuation state $\delta_{a,t}$. 
         \item The control loop starts again from step 1. until convergence, i.e. until an accuracy or stability criterion is satisfied. 
\end{enumerate} 

\begin{table}[!h]
\centering
   \caption{Data acquisition parameters for each robot. $n_l$ sample points resulting in $n_s$ samples per cables. Cables displacement are sampled in $[s_{a, min}, s_{a, max}]$.}
    \centering
    \begin{tabular}{| l || c | c | c | c| }
        \hline
         Robot & $n_l$ & \multicolumn{3}{|c|}{Cables} \\ \hline
         & & $s_{a, min}$ & $s_{a, max}$ & $n_s$\\ \hline
         \textit{Diamond} & 6561 & 0 & 30 & 11\\  \hline
    	 \textit{Finger} & 15000 & 0 & 20 & 81\\  \hline
    \end{tabular}

    \label{tab:parameter}
\end{table}

\section{VALIDATION AND DEMONSTRATION}
\label{VALIDATION}

\subsection{Robots}
\label{robotsDes}

\begin{figure}[!ht]
\centering
\resizebox{.4\textwidth}{!}{
\includegraphics{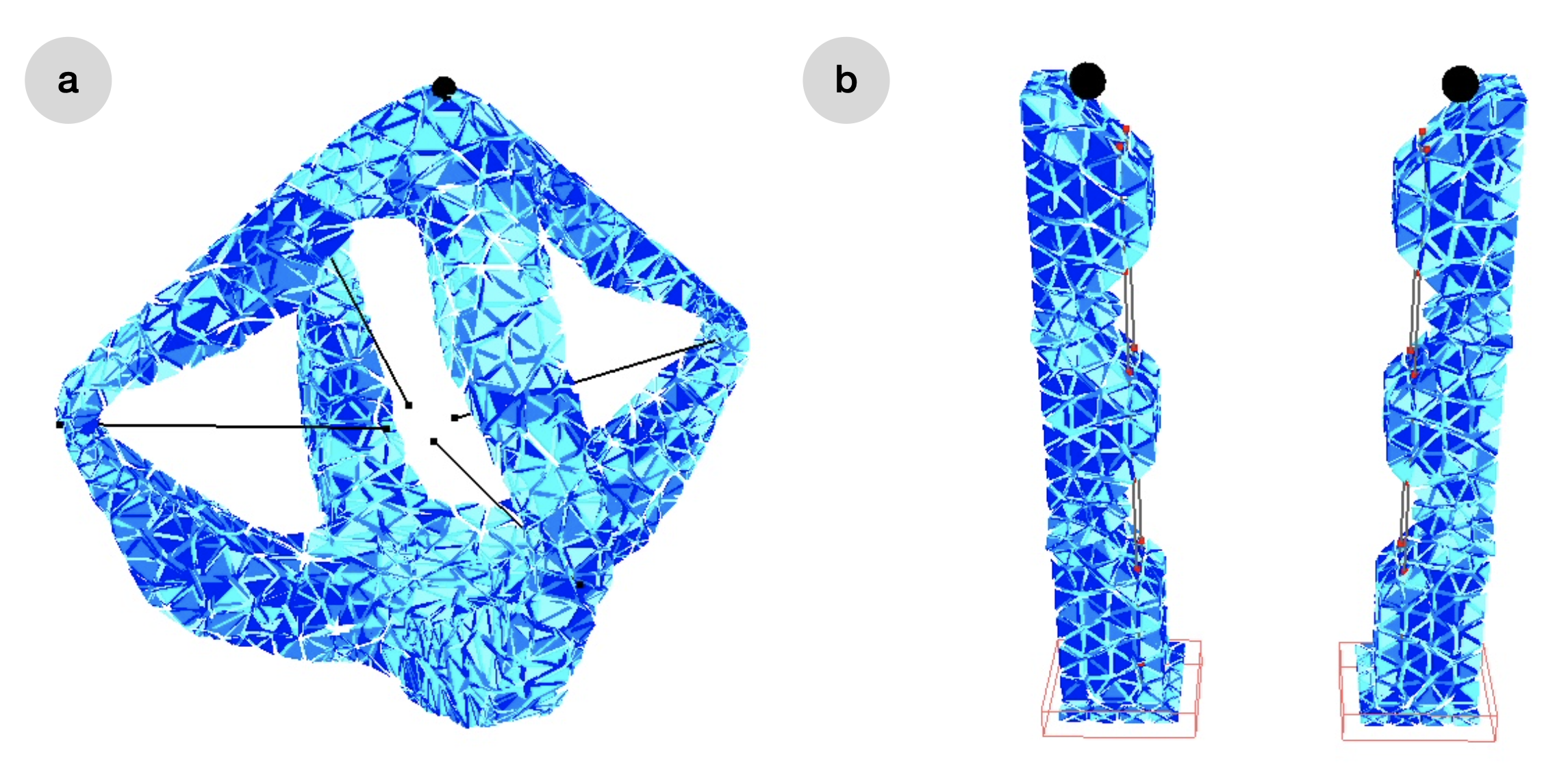}
}
\caption{SOFA simulation of the two robots considered in this study. 3D simulation meshes and effectors are shown respectively in blue and as black dots. a) The Diamond robot controlled in position using an effector located on its tip. b) The Gripper robot controlled using effectors located at the tip of each finger.}
\label{robots}
\end{figure}

The method is illustrated and validated using two FEM models of very different soft robots driven by cables. Although the method can be applied to other types of actuation, cable actuators have been chosen for showing that the unilateral aspect of this type of actuation can be managed. Indeed a cable can be pulled but not pushed. The two used robots are introduced Fig. \ref{robots} and described below: 

\begin{itemize}
    \item The Diamond robot is a soft parallel silicone robot, driven by 4 cables. The FEM model of this robot has been validated in \cite{duriez2013control}.
    \item The Gripper robot uses two identical silicone Fingers, validated in \cite{navarro2020model}. They are made up of three segments connected by accordion-shaped joints and actuated by 2 cables. As this robot is made of two identical Fingers, the mechanical matrices of the system can be built from the matrices computed on a single Finger.
\end{itemize}

These two robots are simulated in SOFA \cite{SOFA} using components from SOFA plugins. The main interest of these robots is that it is possible to physically build them and control them from the simulation using an inverse controller. Simulation results act here as ground truth, allowing both to easily retrieve training data and to apply control schemes. The transition from FEM simulation to real robots has already been addressed in above mentioned works. 
 
For each robot, the data are retrieved from the simulation as described in \ref{pipeline}. Data acquisition parameters for each robot are summarized in Table \ref{tab:parameter}. In these examples, the actuators considered are cables. During data acquisition, the space of displacements i.e. the space of relative lengths $s_{a,t}$ imposed on the cables is discretized and the actuation state $\delta_{a,t}$ is observed as the reached length of the cables at equilibrium. Note that many more samples are considered for the Finger than for the Diamond. Indeed, since more extreme positions and then more deformations are considered when the Finger closes on itself, more data are required for capturing larger evolutions of the learned mechanical features.

\begin{figure*}[!ht]
\centering
\resizebox{0.85\textwidth}{!}{
\includegraphics{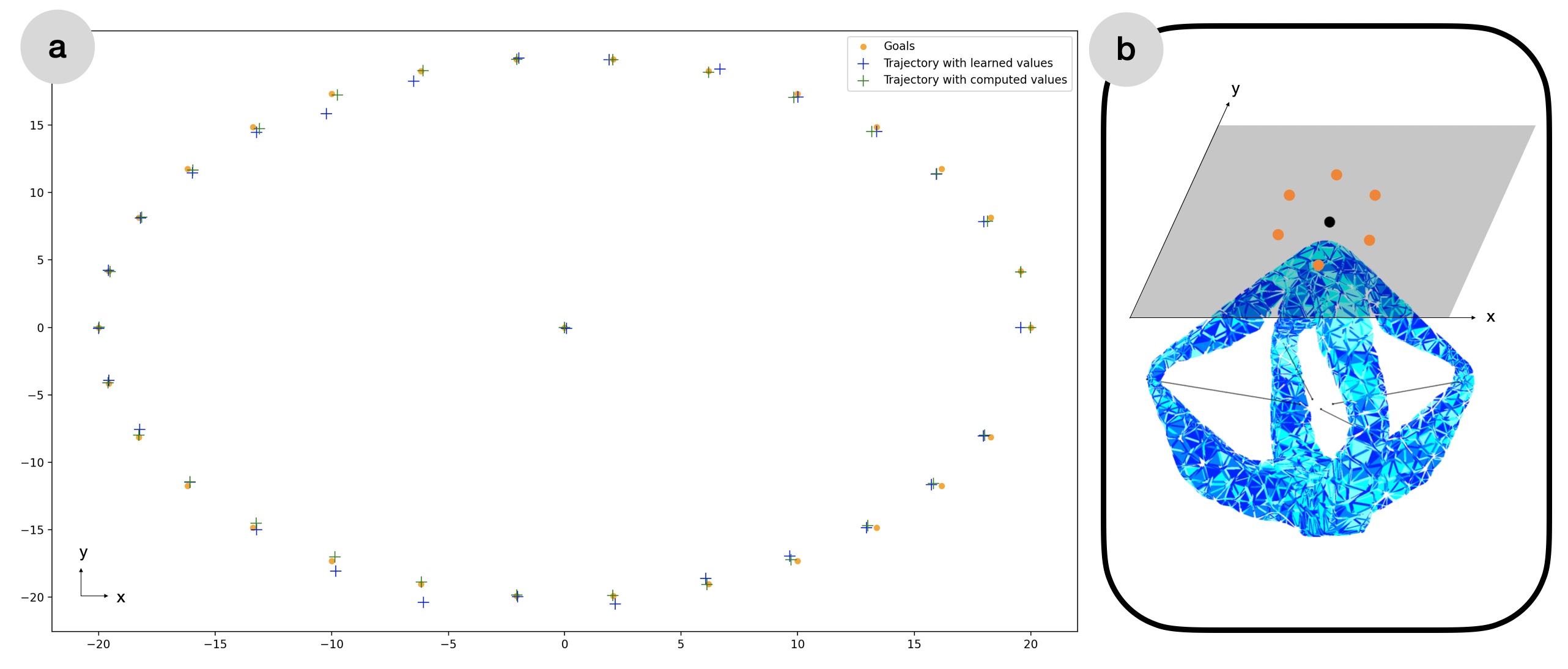}
}
\caption{Control results for the Diamond. a) Trajectory of the end effector in the $(x,y)$ plane for 30 different goals (orange dots) when using learned mechanical matrices (blue cross) or computed mechanical matrices (green cross). b) Illustration of the trajectory to be performed by the Diamond end effector. The end effector of the Diamond is first put in its initial position represented by a black dot. It must then successively reach different targets distributed in a circle around this initial position as shown by the orange dots.}
\label{trajectory}
\end{figure*}

\subsection{Learning results}

A small MLP of 3 layers with 400 weights to learn is sufficient for learning the MR of each robot. For evaluating the learning process, the average loss is computed on the test set of random robots configurations. Learning results are given in Table \ref{tab:final_loss}.

\begin{table}[h]
    \caption{Initial and final test loss values for Diamond and Finger robots. The best trained network encountered is kept at best epoch.}
    \centering
    \begin{tabular}{| l || c | c | c | }
        \hline
     Robot & Initial Loss & Final Loss & Best Epoch\\ \hline
     \textit{Diamond} & 8.1e-1 & 1.2e-4 & 6600\\  \hline
	 \textit{Finger} & 1.0 & 4.4e-6 & 3730\\  \hline
    \end{tabular}
    \label{tab:final_loss}
\end{table}

Although there is more constraints in the Diamond problem, it takes much less samples data for training the MLP for the Diamond than for the Finger. This is because the projected compliance matrix of the Finger robot varies much more than that of the Diamond robot. For the Finger, the influence of the actuators on the end effector changes a lot depending on the configuration of the robot, i.e. depending on how bent the robot is. On the contrary, the effector always has similar behavior in terms of constraints for the Diamond. 

\subsection{Inverse modeling results}

In order to evaluate the use of the learned MRs for inverse modeling, a trajectory is considered in the form of one or several goal positions to be reached by the effector for each robot.As the soft robot dynamics is neglected in this study, the results are only valid for trajectories time-scales in the quasi-static regime. \newline 
 
\subsubsection{Diamond robot control}

A circular trajectory of the effector in a horizontal plane is considered for the Diamond. This trajectory is chosen for being convenient for two dimensions visualization although the robot end effector can be controlled in all three directions of space. Then the positions obtained for each goal using our MR learning based control pipeline and the classical inverse model are compared. Control results are shown in Fig. \ref{trajectory}.

In most configurations the positions reached by the tips when the compact learned model is used is very similar to the one when the full FEM model is considered. This shows that the compact model is a good representation of our system. On some positions, the error between the learned model and the full model are a bit larger: this comes from the learning errors specific to the method. This error can be further decreased by increasing the discretization of the actuation space during data collection and by judicious choices of networks hyperparameters. This example validate the hypothesis that we can use learned quantities $(W_t, \delta^{free}_t)$ for performing control tasks.  

\subsubsection{Gripper robot control}
 
\begin{figure}[!ht]
\centering
\resizebox{0.25\textwidth}{!}{
\includegraphics{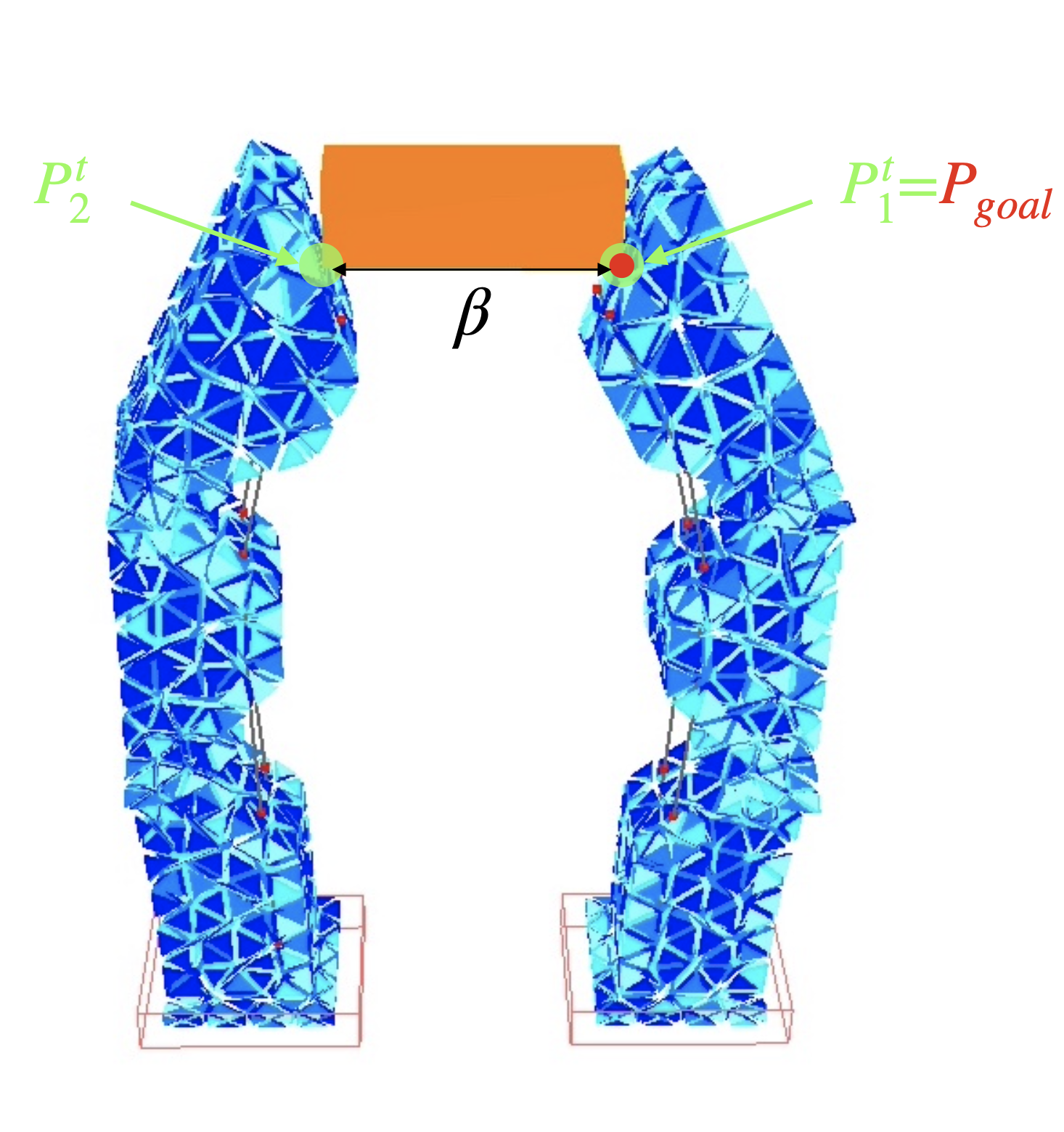}
}
\caption{Position obtained using mechanical coupling between Fingers $1$ and $2$ and learned Finger MR. The position of the effector of Finger $1$ $P_{e_1}^t$ is given by a goal point while the position of the effector of Finger $2$ $P_{e_2}^t$ is imposed by the coupling. This coupling is represented in the picture by an orange object to grasp.}
\label{coupling}
\end{figure}

The training done for a single Finger robot, as shown in Fig. \ref{robots}.b., can be transferred to the Gripper robot. The learned MR of the two Fingers are mechanically coupled by adding a force between the effectors. This force corresponds to a configuration in which the two Fingers catch an object by enabling one Finger to reach a position while forcing the other Finger to follow it. To take this coupling into account, the optimization problem from Eq. \ref{eq:QPActuation} is reformulated according to: 

\begin{equation}
    \begin{aligned}
     arg_{\lambda_{a,1}^t, \lambda_{a,2}^t, \lambda_e^t}&min \quad ||P_{e_1}^t-P_{goal}||^2 \\
    s.t. &  P_{e_1}^t + \beta = P_{e_2}^t \\ 
    \end{aligned}
\label{eq:QPCoupling}
\end{equation}

where $P_{e_1}^t$ and $P_{e_2}^t$ are the respective positions of the effectors of Fingers $1$ and $2$ at time $t$, $P_{goal}$ the target position of $P_{e_1}^t$, $\beta$ a distance describing the size of the object, $\lambda_{a,1}^t$ and $\lambda_{a,2}^t$ the forces applied on the cable actuators of respectively Fingers $1$ and $2$ at time $t$ and $\lambda_{e}^t$ the coupling force shared on the two Fingers. Both the quantities $P_{e_1}^t$ and $P_{e_2}^t$ can be written with the compliance matrices projected in the constraint space as 

\begin{equation}
\begin{aligned}
P_{e_1}^t = P_{e_1}^{t-1} + \widetilde W_{ea,1}^t (\lambda_{a,1}^t -\lambda_{a,1}^{t-1}) +  \widetilde W_{ee,1}^t (\lambda_{e}^t -\lambda_{e,1}^{t-1}) \\
P_{e_2}^t = P_{e_2}^{t-1} + \widetilde W_{ea,2}^t (\lambda_{a,2}^t - \lambda_{a,2}^{t-1}) -  \widetilde W_{ee,2}^t (\lambda_{e}^t -\lambda_{e}^{t-1}) \\
 \end{aligned}
\label{eq:Pi}
\end{equation}
where the $\widetilde W_{ea}$ and $\widetilde W_{ee}$ values are the predicted projected compliance values. The quantities at time $t-1$ are known quantities that are not involved in the optimization. 

This optimization problem enables to mechanically couple the two Fingers using the values learned for a single Finger. Indeed it is possible to build $\widetilde W_2$ from the same network used to predict $\widetilde W_1$ by paying attention to the orientation of the Fingers. Solving this optimization problem applied to the Gripper robot enables to catch an object. Starting from the position given in the Fig. \ref{robots}.b, the resolution of Eq. \ref{eq:QPCoupling} leads to the result given in the Fig. \ref{coupling}. The movement of the gripped object between the two Fingers can be controlled. However, there is a limitation when the force applied to the effectors of the Fingers becomes important for extreme actuation positions. To solve this problem, it would be necessary to integrate the possibility of putting forces on the effector during the learning stage.

\section{DISCUSSION, PERSPECTIVES AND CONCLUSION}
\label{CONCLUSION}

In this paper we introduced a method for representing a soft robot in a compact way. This representation takes the form of MR learning representing the quantities $W_t$ (compliance matrix projected in the constraint space) and $\delta_t^{free}$ (free value of the constraints) given the constraint state $\delta_{a,t}$ and the initial value $W_0$ and $\delta^{free}_0$. The particularity of this method is that these quantities are learned from a finite set of robot configurations. We showed that this method enables for controlling robot with quadratic optimisation. Moreover, since prediction is made using a low-dimensional network, it is very suitable for embedding the control algorithm on a microprocessor where it is not possible to fully simulate the FEM model. Although the method has been illustrated on soft cable robots, nothing prevents it to be used with a wider variety of actuators such as pneumatic chambers or for rigid robots. In addition, even though the framework has been introduced with the quasi-static assumption, it can be extended to soft robot dynamics by taking into account mass and damping matrices in the expression of $W$. 

The model has currently some limitations. First, we do not take into account the contacts between the environment and the robot. Each contact point represents a new constraint that must be integrated into the input data. MLP networks can not handle dynamic resizing of the input and output data. We will test Graph Neural Network for learning a model independent of the number of constraints and directly integrate relational biases in its data structure. As described in \cite{GNN}, message passing could be used to incrementally update the attributes of the graph on which we learn the matrix. In the MR learning case, the value  $W_{ij}$ in $W$ are the influence of the constraint $i$ on the constraint $j$. These values therefore represent relationships between quantities, and it seems natural to us to represent these data in the form of graphs where the nodes $v_i$ represent the constraints defined by its value $(\delta_{a}, \delta^{free})_i$ and the edges $e_{ij} = W_{ij}$ are the interaction between constraint $i$ and constraint $j$. This graph could be initialized with  $W_0$ and $\delta^{free}_0$, and the update with message passing to predict $W_t$ and $\delta^{free}_t$. We believe that this model will enable handling the contacts in future works.

Secondly, although some research works are interested in creating a reduced model of a soft robot \cite{MOR}, actuation space scaling is usually the blocking factor. Indeed collecting data by discretizing the actuation space is exponential in the number of actuators and becomes very quickly impractical. For managing cases with more actuators, a discretization of the effector space could be considered when the number of actuators increase. We also showed with the example of the Gripper robot that individually learned model could be combined for building the model of a larger robot. Thus another solution when we can obtain very distinct modules constituting a robot is to learn separately their MR.

Finally, this approach has the particularity of providing a differentiable function $F$ linking the output of the network $W_t$ and $\delta_t^{free}$ with its input, i.e. with the actuation state $\delta_{a,t}$ of the robot or the initial compliance matrix projected in the constraint space $W_0$. The creation of a design loss from $W_t$ and/or $\delta_t^{free}$ would allow to directly optimize the robot design without the need to simulate it. This is a critical point in design optimization loop when many designs have to be evaluated. The creation of a control loss would also allow to find suitable actuation efforts for a task to be performed. Finally, since the network represents the whole robot in a compact way and since it enables to predict the evolution of this representation according to the robot's actuation state, it is possible to use it in a model based Reinforcement Learning approach for learning high level control tasks. 

\section{ACKNOWLEDGMENT}
Tanguy Navez would like to acknowledge the support of both the Region Hauts de France and the ANR through the program AI\_PhD@Lille co-funded by the University of Lille and INRIA Lille - Nord Europe.

\addtolength{\textheight}{-12cm}   





\end{document}